\documentclass[11pt]{article}

\usepackage[preprint]{acl}

\usepackage{times}
\usepackage{latexsym}

\usepackage{graphicx} 

\usepackage{array}
\usepackage{booktabs}
\usepackage{float}
\usepackage{caption}
\usepackage{cite}
\usepackage{amsmath}

\usepackage{graphicx}
\usepackage{enumerate}
\usepackage{array}

\usepackage{multirow}
\usepackage{comment}
\usepackage{mathtools}

\usepackage{amsfonts}
\usepackage{url}
\usepackage{amsthm}
\usepackage{threeparttable}
\usepackage{arydshln}
\usepackage{booktabs}

\makeatletter
\newcommand{\customdash}[3][0.5ex]{
  \noalign{\ifnum0=`}\fi
  \hrule height #1
  \dashfill{#2}{#1}{#3}
  \futurelet\@let@token\@xhline
}
\makeatother

\usepackage{algorithm}
\usepackage{algpseudocode}
\usepackage{tikz}

\usepackage{booktabs} 
\usepackage{tabularx}
\usepackage{tcolorbox}
\usepackage{natbib}

\usepackage[T1]{fontenc}

\usepackage[utf8]{inputenc}
\usepackage{dashrule}
\usepackage{microtype}

\usepackage{inconsolata}
\usepackage{enumitem}
\usepackage{graphicx}

\author{
  \textbf{Renlong Jie\textsuperscript{1}},
  \textbf{Che Chu\textsuperscript{2}},
  \textbf{Zhen Wang\textsuperscript{3}}
\\
\textsuperscript{1}
Northwestern Polytechnical University,\\
\textsuperscript{2}
School of Statistics and Mathematics,
Yunnan University of Finance and Economics,\\
\textsuperscript{3}iOPEN,
Northwestern Polytechnical University
}

\title{Capability-Aware Early-Stage Research Idea Evaluation}

\begin{document}

\maketitle

\begin{abstract}
Predicting the outcomes of research ideas at their conceptual stage (i.e. before significant resources are committed) holds great potential for optimizing scientific resource allocation and research planning. While existing methods rely heavily on finished manuscripts or peer reviews, we propose a novel capability-aware framework that predicts paper acceptance and ratings using only author information and research ideas, without requiring full text or experimental results. Our approach integrates author information, (inferred) capability presentation, and research ideas through a three-way transformer architecture with flexible fusion mechanisms. We also introduce a two-stage architecture for learning the capability representation given the author information and idea. Experiments show that our method significantly outperform the single-way models by finetuning bert-base and bert-large, and the capability predicting significantly increase the predictive accuracy of the final model. The proposed method can be applied in both early-stage research outcome prediction and scientific resource allocation.
\end{abstract}



\section{Introduction}

Research quality evaluation and analysis is an important topic in science of science \citep{fortunato2018science, merton1968matthew}. Generally, the success of a research depends not only on the research ideas but also on the capability of the collaborative researchers and collaboration quality \citep{tigges2019measuring}. The goodness of the research idea or novelty is shape the upper bound of the paper impact and the venue it can be published. Meanwhile, the capability of the the authors, especially the first/co-first authors and the corresponding authors, can determine both the methodological rigor and the overall presentation quality of a research work. Merely having good ideas without the matching capability still cannot produce high-quality work. Beyond that, the identity of authors can also make a significant difference of paper on paper acceptance due to their networking and resources \citep{lee2013bias, kozlowski2022intersectional, sutton2023patrick}.

\begin{figure}[t]
\begin{center}
 \includegraphics[width=1.0\linewidth]{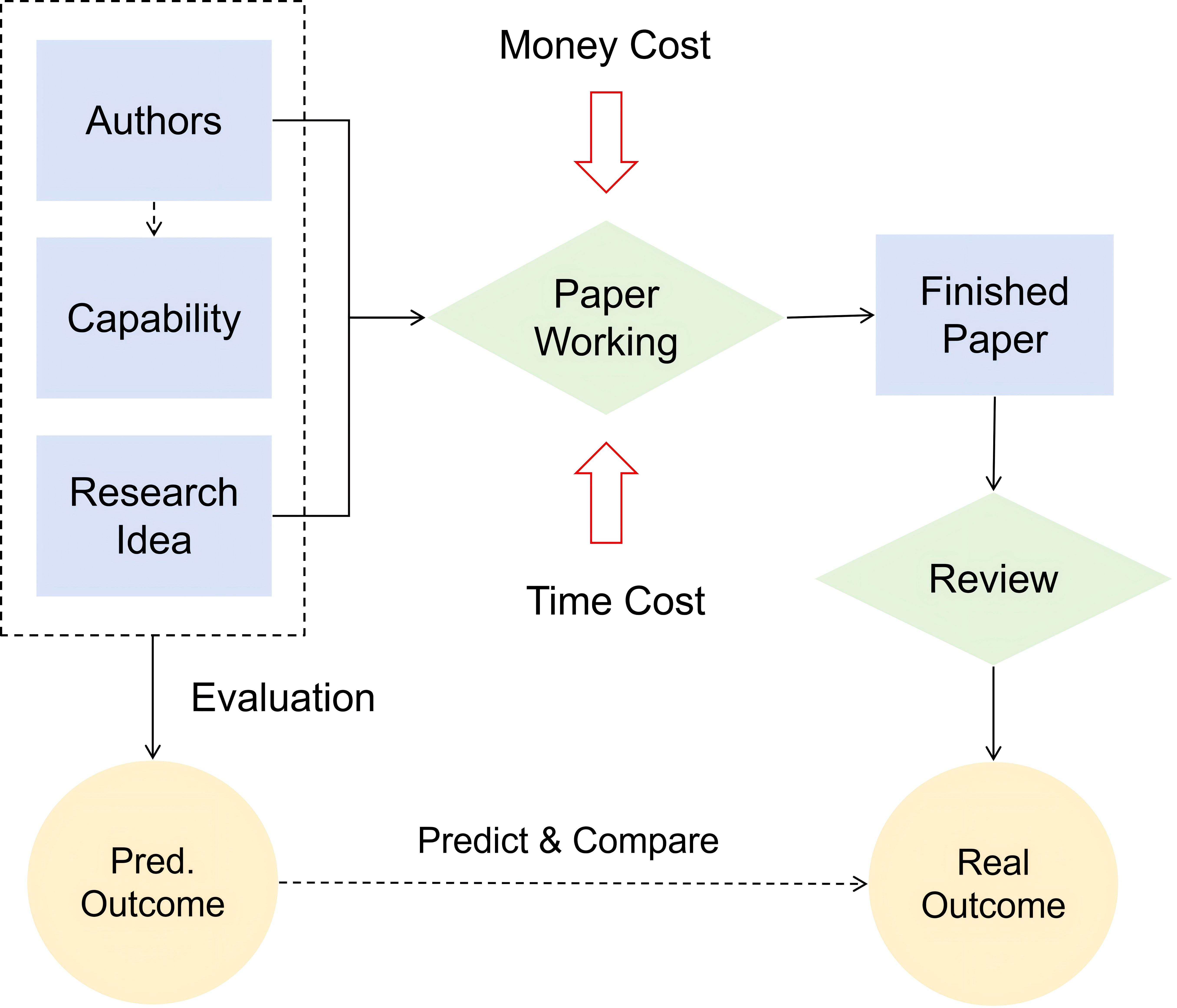}
 \caption{Diagram of author identity\&capability aware Research Idea evaluation.} \label{Fig:Concept}
\end{center}
\end{figure}

With the development of large language models (LLMs), many researchers try to generate research ideas with them and find their potential to develop highly innovative and applicable ideas \citep{li2024chain, hu2024nova, pu2025ideasynth, sican2025, baek2025researchagent, su2024many}. Given a research idea, the cost of a research work can still be large in both time and funding, especially in the fields that involve data collection or large-scale computation \citep{vivona2023costs, cottier2024rising}. However, most of the existing work for idea evaluation and acceptance prediction are based on pre-review finished papers or reviewers' comments \citep{zhao2025review, Feng2025GraphEval, thelwall2025evaluating}. Therefore, it is necessary to develop methods that can evaluate the quality of research idea and whether the research idea match the capability of a research group. Ideally, this should be done during the research design or at the very beginning of a research project.

 
 In this study, we aims to build effective predictive models to evaluate if a research idea is promising to achieve a good research outcome in top AI conferences such as ICLR/NeurIPS, given a group of researchers. For this, we also develop a method for infer the capability of the authors when they working on a research idea. We collect the author information and research ideas from more than 20,000 papers from top machine learning conferences published between 2023 and 2025. We extract the inferred author capabilities with large language models. The main contributions include the following aspects:
\begin{itemize}[nosep, leftmargin=*, label=\textbullet, align=left, itemindent=0pt]
\item We apply LLM to extract the main research ideas and infer the capability of author groups for all open-source submissions in NeurIPS 2023/2024 and ICLR 2024/2025.
\item We investigate the impacts of author information, author-group capability and research ideas on the review outcomes in terms of average rating and acceptance decision.
\item We built a model that can effectively predict the acceptance and rating of top-tier machine learning conferences.
\item We develop a two-level architecture for predicting capability representation with author information.
\end{itemize}

\section{Related Work}

\subsection{Research outcome prediction}

Existing work on paper acceptance prediction predominantly relies on analyzing finished manuscripts with known experimental results. For instance, \citet{kang2018dataset} introduced the PeerRead v1 dataset, containing paper drafts and peer reviews, which has been widely used for acceptance prediction and review scoring. \citet{wang2018sentiment} introduces a multiple instance learning network with an abstract-based memory mechanism (MILAM) to predict overall recommendations (accept/reject) and identify sentiment polarities in peer reviews. \citet{ghosal2019deepsentipeer} explores leveraging reviewers' sentiment embedded in peer review texts to predict manuscript acceptance or rejection, proposing a deep neural architecture that integrates paper content, reviews, and sentiment polarity. \citet{skorikov2020machine} utilize machine learning method to predict whether a scientific paper will be accepted in a top-tier AI conferences or not. \citet{kang2022automatic} introduces a modularized hierarchical attention network (MHAN) for automatic academic paper rating (AAPR), achieving an accuracy of 65.33\% for predicting the acceptance decision. \citet{rimkus2023reviewer} addresses the subjectivity in peer review by developing and evaluating transformer models to predict paper acceptance. \citet{Feng2025GraphEval} proposed GraphEval, a graph-based framework that evaluates ideas via viewpoint decomposition and novelty detection. Overall, existing works on research manuscript/idea evaluation are based on the finished paper with known experimental results.

\subsection{LLM-based paper analysis}

In recent years, many studies use LLMs to evaluating and extracting key information from scientific papers. \citet{newman2024arxivdigestables} proposes a decomposition-based framework using language models to automatically generate comparative tables from scientific papers. \citet{csahinucc2024efficient} proposes an LLM-based framework to automatically extract and normalize Task-Dataset-Metric-Result tuples from papers, enabling leaderboard construction in both predefined and cold-start settings. \citet{xu2025can} introduce a benchmark for evaluating LLMs’ ability to identify scientific limitations, combining synthetic perturbations and real human critiques. Other works are dedicated to building an automatic review system for generating full paper reviews \citep{zeng2023meta, jin2024agentreview, staudinger2024analysis, kuznetsov2024can, d2024marg}. 

\begin{figure*}[th]
\begin{center}
 \includegraphics[width=0.85\linewidth]{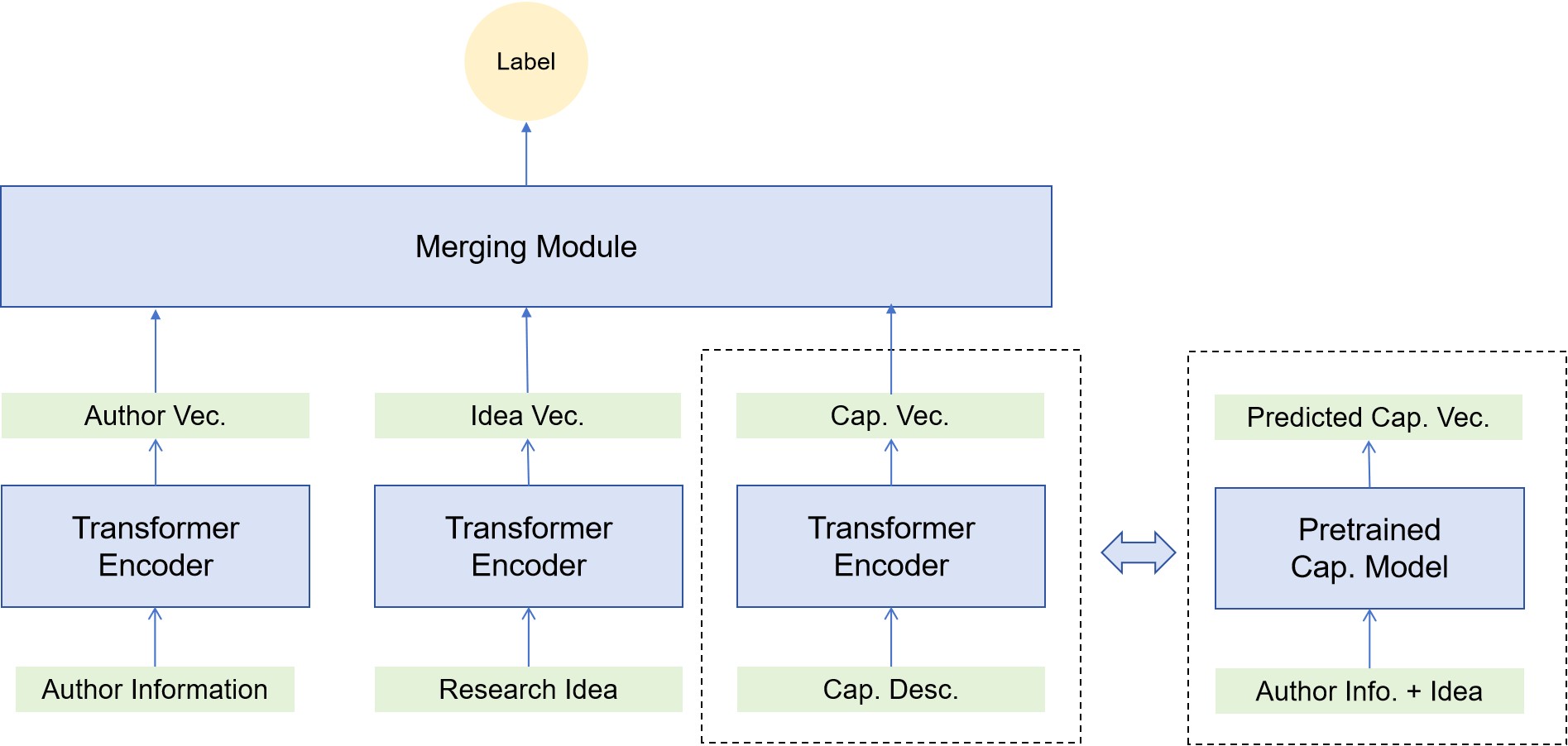}
 \caption{Diagram of review outcome prediction model. Three independent transformer encoder models are applied for processing the author information, research idea and capability, respectively. A merging module is applied to merge the three top vectors for predicting acceptance/ratings. The capability model can be replaced by the capability representation prediction model for the scenario when capability description is not explicitly given.} \label{Fig:S2}
\end{center}
\end{figure*}

\section{Task Description}


This study aims to develop a model for predicting the acceptance and ratings of research papers using only author information and research ideas (without full texts or results) to forecast outcomes before major effort is invested. The tasks include: (a) Constructing a dataset with author profiles, extracted capabilities, and research ideas derived from papers using carefully designed prompts. (b) Proposing a transformer-based model to predict acceptance and average ratings based on the target venue, author information, research idea, and author capabilities. (c) Inferring author capabilities from their existing works and predicting review outcomes using only author information and research ideas. 

The proposed approach is expected to enable early-stage prediction of paper acceptance and supports optimal matching between author groups and research ideas to maximize research quality. 

\section{Method}

\subsection{General architecture}

We propose a joint architecture (Figure~\ref{Fig:S2}) to assess research ideas and author-idea fit for paper quality prediction. The model encodes author information, capability, and research ideas into separate representations, then merges them to predict review outcomes, which is end-to-end trainable. Unlike single-encoder transformers, our approach better captures interactions between these distinct textual components, similar to user-item embedding interactions in recommender systems \citep{he2017neural, chen2019joint, pugoy2020bert}. Additionally, we train an independent module to predict capability representations from author information, enabling applications where only author identities, ideas, and historical research outputs of different author groups are available.


\subsection{Feature Fusion Strategies}
\label{sec:fusion}

To predict review outcomes from author information, capability representations, and research ideas, we demonstrate two efficient fusion strategies for merging the top representations from transformer encoders. Other investigated fusion strategies are discussed in Appendix~\ref{Sec:A.1}.

\subsubsection{Self-Attention Mechanisms (SA1, SA2)}
\label{subsec:sa1}

We employs a self-attention mechanism to model interactions between feature vectors \citep{de2022attention}. Given three feature representations $\mathbf{h}_{\text{author}}$, $\mathbf{h}_{\text{cap}}$, $\mathbf{h}_{\text{idea}} \in \mathbb{R}^d$, we first stack them into a sequence matrix:

\begin{equation*}
\mathbf{X} = \text{stack}([\mathbf{h}_{\text{author}}, \mathbf{h}_{\text{cap}}, \mathbf{h}_{\text{idea}}]) \in \mathbb{R}^{3 \times d}
\end{equation*}
We then compute the self-attention mechanism:

\begin{align*}
\mathbf{Q} &= \mathbf{W}_q \mathbf{X},\quad 
\mathbf{K} = \mathbf{W}_k \mathbf{X},\quad 
\mathbf{V} = \mathbf{W}_v \mathbf{X} \\
\mathbf{Z} &= \text{softmax}\left(\frac{\mathbf{Q} \mathbf{K}^\top}{\sqrt{d}}\right) \mathbf{V} \\
\mathbf{Y} &= \text{LayerNorm}(\mathbf{X} + \text{Dropout}(\mathbf{Z}))
\end{align*}
The output is aggregated as $\mathbf{y}_{\text{sum}} = \sum_{i=1}^{3} \mathbf{Y}_i$ and projected to the final prediction $y = \mathbf{w}^\top \mathbf{y}_{\text{sum}}$. Residual connections preserve original features while enabling contextual learning, with layer normalization and dropout ensuring training stability. As this setting is mainly designed for deep self-attention architectures, we further denote the setting with and without residual connection, dropout and layer norm as ``SA2'' and ``SA1'' for ablation study, respectively. 
As self-attention is one major component of transformer layers, we also apply transformer encoder as a fusion strategy (denoted as ``TF'') \citep{vaswani2017attention}, where $\mathbf{X}$ is the input embeddings without positional encoding, while multi-head setting and a MLP layer with add \& norm operation is included. In that case, the third top embedding (on top of idea embedding) is selected for prediction with an extra linear layer. 

\subsubsection{Residual Linear Fusion (R1)}
\label{subsec:r1}

R1 employs a simple yet effective linear transformation approach with element-wise addition for feature fusion. Each feature is independently projected and combined via element-wise addition:
\begin{align*}
\mathbf{h}_{\text{combined}} = \mathbf{W}_a \mathbf{h}_{\text{author}} + \mathbf{W}_i \mathbf{h}_{\text{idea}} + \mathbf{W}_c \mathbf{h}_{\text{cap}},
\end{align*}
where $\mathbf{W}_a, \mathbf{W}_i, \mathbf{W}_c\in \mathbb{R}^{d\times d}$ are learnable parameters. The final output is computed as $y = \mathbf{w}^\top \mathbf{h}_{\text{combined}}$ with $\mathbf{w}\in \mathbb{R}^{d\times 1}$. This provides computational efficiency with linear complexity, projecting features into a shared space where additive fusion effectively integrates information despite lacking explicit interaction modeling.

\subsection{Learning for capability representation} \label{Sec:4.2}

Given the author information, including the authors' names, positions, gender, country, and affiliation, the capability of the author group can be determined. However, in practice, their capabilities can only be estimated by themselves or be inferred from their research outputs. Therefore, we build a method for predicting the capabilities shown in the finished paper with the author information and research idea. As different ideas require different set of capabilities, we assume that author-group capability shown in a research work can be determined by both the author identity information and the research ideas they work on.

We assume each researcher has a particular capability vector, which is denoted as $\mathbf{c}_i$, $i\in {1,...,N}$. The group capability shown in a particular paper $j$ is denoted by $\mathbf{c}_{group,j}$, which is determined by the both the capability vectors $\mathbf{c}_1,...,\mathbf{c}_n$ of all the authors in the author-list $a^{text}_{j}$, and the representation of the research idea of the paper $\mathbf{p}^{text}_{j}$. This cis given by:
\begin{equation*}
\begin{split}
\mathbf{c}^{group}_{j} = f(\mathbf{c}_1,...,\mathbf{c}_n; \mathbf{p}^{text}_{j}) =f(\phi(\mathbf{a}^{text}_{j}), \mathbf{p}^{text}_{j}),
\end{split}
\label{Eq:1}
\end{equation*}
where $\phi(.)$, $f(.)$ are two transformer encoder models in this work. The model architecture, illustrated in Figure~\ref{Fig:S1}, involves two stages:

\begin{figure}[t]
\begin{center}
 \includegraphics[width=1.0\linewidth]{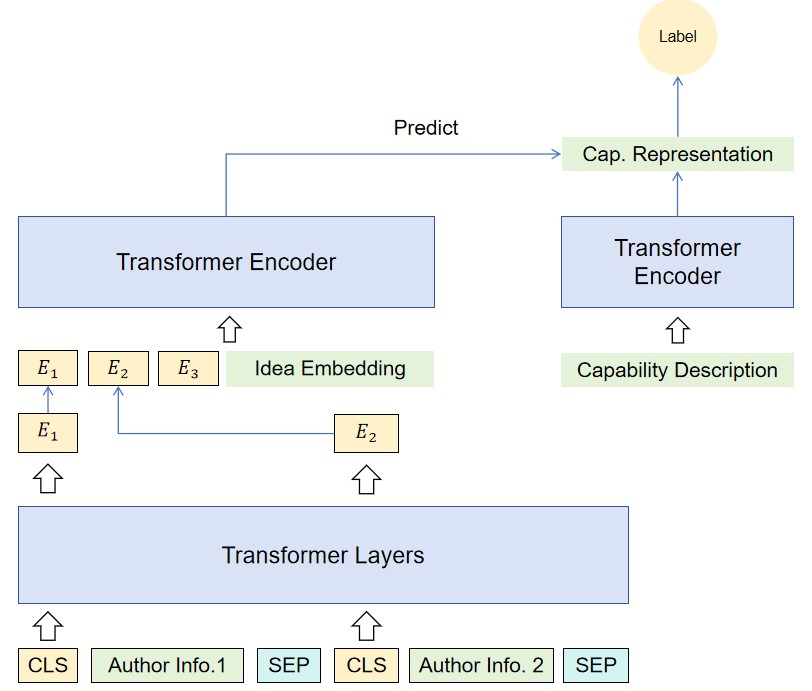}
 \caption{Diagram of the capability representation prediction module. We apply author information and idea to predict the capability representation shown in a given research work.} \label{Fig:S1}
\end{center}
\end{figure}

\textbf{(a) Capability Model Pretraining}: We extract a templated author capability description from each paper and fine-tune a BERT model~\citep{devlin2019bert} to predict paper acceptance or rating decisions. This model provides a capability representation on top of the transformer encoder.

\textbf{(b) Capability Prediction}: With the pre-trained capability model fixed, we train a bi-level transformer model that takes author information $a^{text}_{j}$ and research idea $p^{text}_{j}$ as inputs to predict the capability representation. In the first level, we apply segmentation embeddings on textual information of different authors started with [CLS] tokens (see Figure~\ref{Fig:S1}), the top vectors of [CLS] tokens are considered as the capability representations of authors in the author list. They are concatenated together along with the idea embedding, which is obtained by passing idea tokens through the embedding layer of the second-level transformer encoder. The concatenated representation sequence are feed into the second-level transformer-encoder, whose first output representation is applied for predicting the capability representation.   

To train the capability representation prediction model, we introduce a multi-objective learning mechanism. The loss function is:
\begin{equation}
\begin{split}
L = &-sim(\mathbf{\hat{c}}, \mathbf{c}) + \lambda_1 \text{max}(sim(\mathbf{\hat{c}},\mathbf{a}),0)\\
& + \lambda_2 MSE(y_{\hat{c}}, y_c),
\end{split}
\end{equation}
where $sim(\mathbf{a}, \mathbf{b}) = \frac{\mathbf{a}^T \mathbf{b}}{||\mathbf{a}||||\mathbf{b}||}$ is the cosine similarity, $\mathbf{c}$ and $\mathbf{\hat{c}}$ are the capability representation and the predicted capability representation, respectively. $\mathbf{a}$ is the top vector of a pre-trained author information model (to predict the label) on the same instance. $y_{\hat{c}}$ and $y_c$ are predicted labels or outputs by estimated and original capability representations through a shared linear layer. $\lambda_1$ and $\lambda_2$ are hyper-parameters to be tuned and we set them to be 0.2 and 1, respectively. The first term is to maximize the similarity between the predicted representation and capability representation, and the second term is to minimize the similarity between the predicted representation and the author information representation to avoid multi-collinearity. It pushes the similarity between predicted cap. representation and author representation to 0.

\subsection{Prompt design}\label{Sec:4.4}

We apply prompts to collect author capability and research ideas in standardized formats without involving post-research information. The prompt for extracting the research idea is given in Prompt 1, where we extract the research problem, the method applied in the paper to solve the problem, and the innovation involved in the method. Different from the abstract as applied in \citet{Feng2025GraphEval}, we do not include the experimental outcome in the research idea, which ensures that it can be evaluated in the early stage or during research design.
\begin{figure}[th]
\begin{center}
 \includegraphics[width=1.0\linewidth]{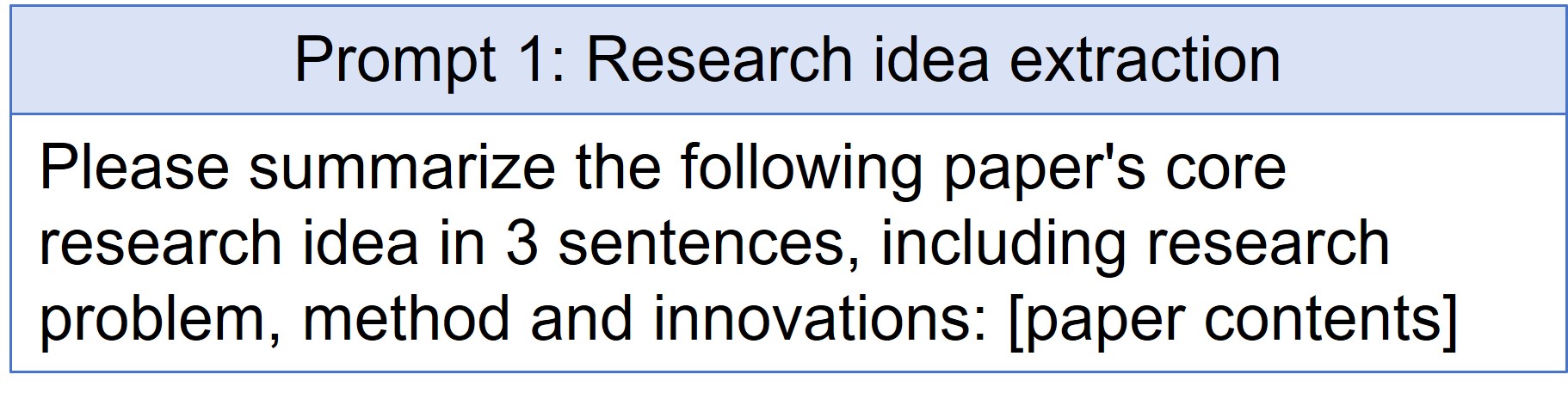}
\end{center}
\end{figure}

The prompt for capability extraction is given in Prompt 2. We apply a template to extract the authors' skills and proficiency, including six general skills: mathematical derivation, theoretical analysis/proving, model/architecture design, data collection and experimental design, and 5-10 expertise that particularly owned by the author group. Meanwhile, we extract the computing in GPU types/hours, financial and time budgets as we believe computing power and financial budget can also be considered as capability of an author group, which can also be planned in research design.
\begin{figure*}[th]
\begin{center}
 \includegraphics[width=0.90\linewidth]{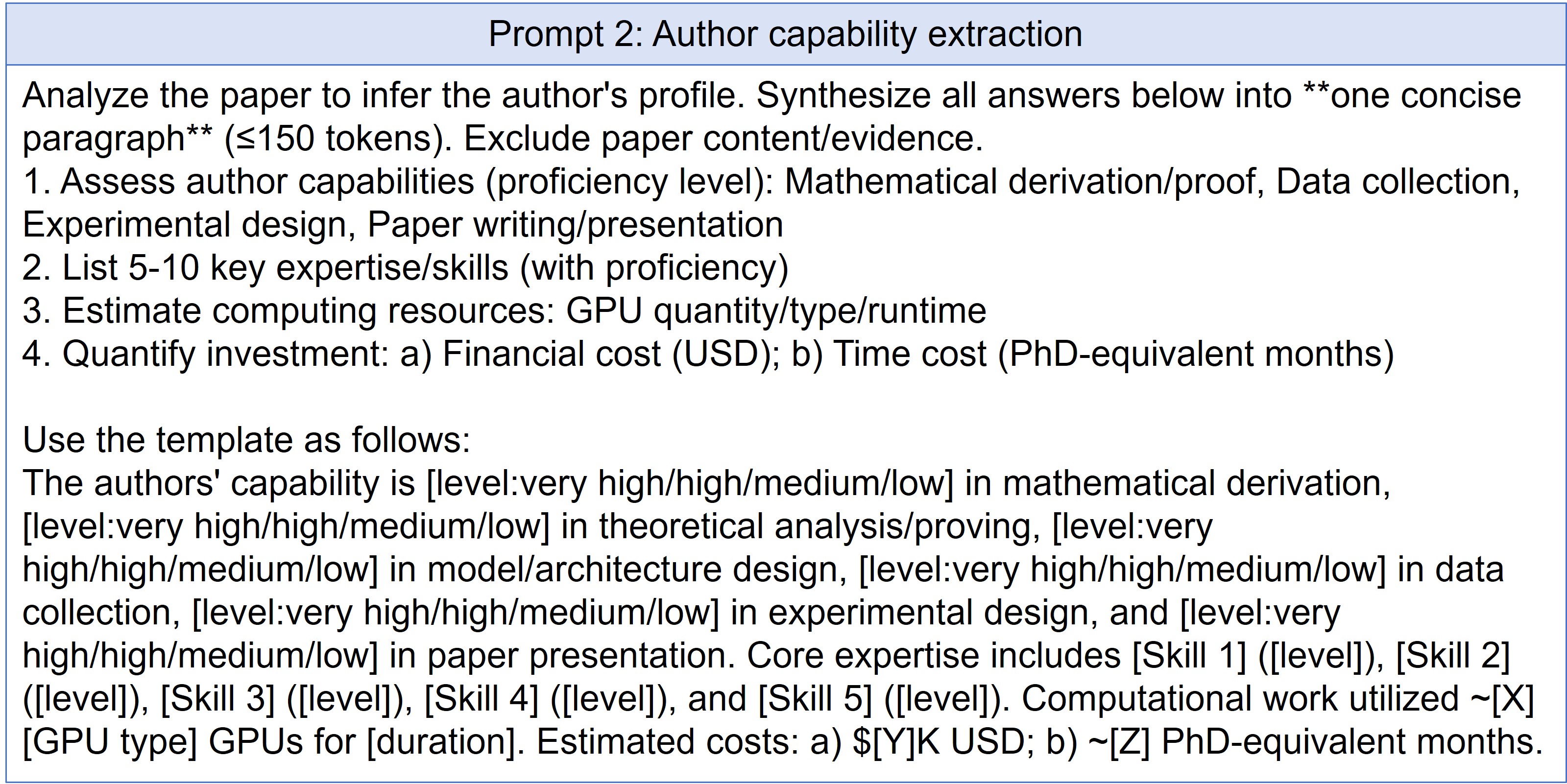}
\end{center}
\end{figure*}

\section{Experiment}

\subsection{Setting up}  \label{Sec:5.1}

We download the pdf files of 25,889 submission from Openreview, including the submission of ICLR 2024/2025 and NeurIPS 2023/2024 that are open-source. 
We merge the papers from this two conferences as their acceptance bars are close to each other.
As our method involves the training of author capabilities, we selected about 16,712 papers where the first author appears twice in all the papers.

We parse the pdf files with grobid \citep{lopez2009grobid}, and use Deepseek-R1\citep{bi2024deepseek, guo2025deepseek} to extract the research ideas and author capabilities in pre-defined formats (see Section~\ref{Sec:4.4}). Meanwhile, we directly collect the author information, title, abstract, ratings and acceptance decisions from the Openreview API. We also apply venue information (year and conference) for average rating prediction. For acceptance prediction, we do not apply venue information as the available open-source NeurIPS papers are mostly accepted papers. 

We shuffle the combined dataset with a random seed of 42, and further split it into training set, validation set and test set with a ratio of 8:1:1. That is first 80\% of the shuffled dataset is selected as the training set, the next 10\% as the validation set, and the last 10\% as the test set. For each experiment, we perform five trials using different random seeds from $\{42,0,1,2,3\}$ to initialize the model parameters (if not pre-trained) as well as the dataloader, select the model checkpoint with the minimum validation loss, and report the medium or mean \& standard deviation of the corresponding model performance on the test set. Experiments are done with 2 RTX 4090 GPUs in the environment of pytorch 2.5.1+transformers 4.54.1.

\subsection{Transformer encoder baseline} \label{Sec:5.2}

We fine-tune a BERT-base model on various input combinations to predict paper acceptance and average ratings \citep{devlin2019bert}: (a) title, (b) title+abstract, (c) author information, (d) capability, (e) idea, (f) author+idea (AI), (g) capability+idea (CI), and (h) author+capability+idea (ACI). Venue information is included in all settings for average rating prediction to account for rating distribution differences across conferences and years. 

We adopt a discriminative fine-tuning strategy: backbone parameters are updated with a learning rate of $2\times 10^{-5}$, while the prediction head uses $1\times 10^{-4}$. Optimization uses AdamW (weight decay 0.01) and a linear warm-up scheduler (10\% of total steps) to ensure stable convergence \citep{kingma2014adam, loshchilov2017decoupled}.

Table~\ref{tab:bert_rating_acc} reports average rating and acceptance prediction performance. For rating prediction, models using capability or combined inputs (e.g., ACI) achieve slightly lower MSE/MAE errors than those using author or idea alone. For acceptance prediction, capability-inclusive settings (e.g., ACI) yield higher accuracy and F1, with BERT-large/ACI reaching 64.55\% accuracy among non-text inputs. Text-based inputs with abstract perform strongest overall (e.g., 68.48\% accuracy for BERT-base), underscoring the complementary role of full text alongside capability features.


\begin{table}[htbp]
  \centering
  \small
  \caption{Summary of finetune BERT performance on different settings for average rating and acceptance prediction (in percentage).}
    \begin{tabular}{ccccccc}
    \toprule
    \multirow{2}{*}{Model} & \multicolumn{2}{c}{Rating} & \multicolumn{4}{c}{Acceptance} \\
    \cmidrule(lr){2-3} \cmidrule(lr){4-7}
     & MSE & MAE & Acc & P & R & F1 \\
    \midrule
    title & 1.171  & 0.852  & 58.8  & 54.5  & 17.0  & 25.5  \\
    title+abs & 1.172  & 0.855  & 68.5  & 70.0  & 46.2  & 55.0  \\
    abs   & 1.170  & 0.855  & 68.3  & 67.7  & 48.5  & 56.3  \\
    \hdashline
    \noalign{\vskip 3pt}
    author & 1.171  & 0.852  & 59.3  & 53.9  & 26.9  & 35.4  \\
    cap   & 1.167  & 0.853  & 64.2  & 62.0  & 39.6  & 48.1  \\
    idea  & 1.170  & 0.853  & 61.3  & 56.9  & 34.6  & 43.0  \\
    AI & 1.169  & 0.852  & 61.5  & 56.0  & 41.9  & 47.8  \\
    CI & 1.169  & 0.852  & 63.6  & 60.5  & 40.1  & 48.0  \\
    ACI & 1.169  & 0.851  & 64.3  & 61.7  & 42.0  & 49.5  \\
    \bottomrule
    \end{tabular}%
  \label{tab:bert_rating_acc}%
\end{table}%

\begin{table}[t]
\centering
\renewcommand{\arraystretch}{0.9}   
\setlength{\tabcolsep}{0pt}        
\small
\caption{Summary of Regression coefficients of model outputs on test set for predicting the average ratings. ***, **, and * indicate 0.001, 0.01, and 0.05 levels of significance, respectively.}
\begin{tabular*}{\linewidth}{@{\extracolsep{\fill}}lcccc}
\toprule
 Model & 1 & 2 & 3 & 4\\
\midrule
Const
 & $-22.90^{***}$ & $-25.41^{***}$ & $-25.27^{***}$ & $-23.41^{***}$\\
 & {(1.998)} & {(2.066)} & {(2.06)} & {(2.00)}\\[2pt]
Author
 & $2.205^{***}$ & $2.778^{**}$ & $3.232^{*}$ & $2.010^{***}$\\
 & {(0.426)} & {(0.628)} & {(0.639)} & {(0.433)}\\[2pt]
Idea
 & $2.992^{***}$ & $1.391^{***}$ & $1.081^{***}$ & $3.095^{***}$\\
 & {(0.568)} & {(0.462)} & {(0.469)} & {(0.568)}\\[2pt]
Capability
 & --- & $1.429^{*}$ & $1.110^{*}$ & ---\\
 & & {(0.666)} & {(0.67)} & \\[2pt]
Cap\_Pred
 & --- & --- & $0.158^{***}$ & $0.190^{*}$\\
 & & & {(0.045)} & {(0.08)}\\
\midrule
R-squared & 0.119 & 0.129 & 0.135 & 0.123\\
\bottomrule
\end{tabular*}

\label{tab:stat_reg}
\end{table}

\subsection{Statistical analysis}

We collect the predicted values of all the transformer encoder models trained on (a)(c)(d)(e) (see Section~\ref{Sec:5.2}), and evaluate their correlation matrix and build a linear regression to fit the average rating. To avoid over-fitting, we fit regression models on test sets. The correlation  matrix are discussed in Appendix ~\ref{Sec:A.2}. For the fitted models in Section~\ref{Sec:5.2}, we collect their predictions on the test set. We use the outputs of different models (trained on different textual information) as the independent variables, and the average rating is selected as the dependent variable. The summary table of OLS is given by Table~\ref{tab:stat_reg}. It shows that capability is always a significant factor after controlling other independent variables. All of author identity, idea and capability show significant impact. This ensures that they should all be included in the joint three-way model. Compared with author identity and idea, the capability is demonstrated to be more significant with a higher slope. Meanwhile, the outputs with predicted capability $Cap.pred$ based on author information is also a significant factor whether the output of capability model exist or not.

\subsection{Top-level merging mechanisms} \label{Sec:4.4}

The choice of top-level fusion mechanism is crucial for the predictive performance of the model. In this experiment, we compare a variety of fusion strategies, including simple averaging/summation, a single-layer Transformer encoder with 1/2/4/8 attention heads (TF-1L-1/2/4/8h), the self-attention mechanism with/without layer normalization and residual connections discussed in Section~\ref{subsec:sa1} (SA1/SA2), the linear fusion mechanism (R1) introduced in Section~\ref{subsec:r1}, and weighted averaging with a gating mechanism. In addition, we use BERT-base and BERT-large models that process the textual information of (author identity, capability, and idea) as a whole as baselines. We also introduce a series of cutting-edge large language models as comparative baselines, including Deepseek V3.2 \citep{liu2025deepseek}, ChatGLM-4.7 \citep{glm2024chatglm}, GPT-4o \citep{hurst2024gpt}, Gemini-3 \citep{team2023gemini}, and Claude-sonnet-4.5 \citep{anthropic2024sonnet45}, to predict the average rating and acceptance probability for each test instance. Detailed setup and prompt templates for LLM-based evaluation are provided in Appendix Appendix~\ref{Sec:E}. Furthermore, we compare the predictive performance when using the abstract (which may contain research outcomes) in place of the pure idea description, to further demonstrate the effectiveness of the fusion models.
\begin{table}[btp]
  \centering{\small
  \caption{Test performance of rating prediction models.}
    \begin{tabular}{ccc}
    \toprule
          & MSE & MAE \\
    \midrule 
    average & 1.233 & 0.885 \\
    \hdashline
    \noalign{\vskip 3pt}
    bert (title+abs) & 1.171 (0.0016)  & 0.855 (0.0037)  \\
    bert-large (title+abs) & 1.172 (0.0023)  & 0.853 (0.0033)  \\
    TF-1l-8h (AC+abs)    & \textbf{0.999 (0.018)} & \textbf{0.776 (0.0075)}  \\
    \hdashline
    \noalign{\vskip 3pt}
    Deepseek-V3.2 & 1.709 & 1.082 \\
    ChatGLM-4.7 & 1.681 & 1.024 \\
    GPT-4o & 1.859 & 1.085 \\
    Gemini-3-preview & 1.573 & 0.977 \\
    Claude-sonnet-4.5 & 1.830 & 1.088 \\
    bert (ACI) & 1.169  (0.0010)  & 0.853 (0.0042)  \\
    bert-large (ACI) & 1.169 (0.0018)  & 0.852 (0.0021)  \\
    R1    & 1.018 (0.0037)  & 0.794 (0.0015)  \\
    Gated Ave.    & 1.024 (0.0181)  & 0.795 (0.0069)  \\
    SA-1 & \textbf{1.013 (0.0045)}  & \textbf{0.789 (0.0024)}  \\
    SA-2 & 1.013 (0.0054)  & 0.794 (0.0018)  \\
    TF-1l-1h & 1.033 (0.0153)  & 0.799 (0.0040)  \\
    TF-1l-8h & 1.036 (0.0113)  & 0.799 (0.0070)  \\
    \bottomrule
    \end{tabular}%
  \label{tab:6}}%
\end{table}%

The results for average rating and acceptance prediction are presented in Table~\ref{tab:6} and Table~\ref{tab:7}, respectively. The tables are partitioned by dashed lines to distinguish average/random baselines, scenarios where the abstract substitutes for the idea, and all configurations involving author information, capability, and idea. Table~\ref{tab:6} compares the performance of various top-level fusion mechanisms for average rating prediction. Our proposed three-way fusion models (SA1, SA2, R1) consistently outperform BERT and BERT-large baselines that process concatenated inputs, achieving lower MSE and MAE. When the paper abstract replaces the pure idea information, the Transformer-based fusion strategy (TF-1L-8h-abs) yields the best results (MSE: 0.9990, MAE: 0.7761). For the extracted idea text, SA1 performs best, demonstrating the effectiveness of explicit interaction modeling between author, capability, and idea representations. The residual linear fusion (R1) also exhibits strong performance, providing a simpler yet effective alternative for feature fusion. Additionally, all large language models applied in a zero-shot manner underperform compared to supervised discriminative models.

Table~\ref{tab:7} presents the performance of different fusion strategies for acceptance prediction. Overall, when the abstract is used as the idea, the 8-head single-layer Transformer encoder as the fusion mechanism achieves the best performance, with 70.45\% accuracy. For the extracted idea text, the self-attention mechanism with residual connections (SA1) achieves the highest accuracy (65.29\%) and precision (64.26\%), surpassing BERT baselines and other fusion methods. The residual linear fusion (R1) also shows strong and balanced performance in precision and recall. These results confirm that structured fusion of multi-source inputs (author, capability, idea) effectively enhances predictive accuracy over single-encoder models, with SA1 providing the most robust fusion scheme for acceptance decision prediction. While zero-shot large language models yield higher recall and F1 scores, their accuracy and precision remain lower than those of supervised discriminative models.

\begin{table}[htbp]
\centering\small
\caption{Test performance of acceptance prediction models (in percentage).}
\begin{tabular}{@{}m{2.8cm}@{}m{1.1cm}@{}m{1.1cm}@{}m{1.1cm}@{}m{1.1cm}@{}}
\toprule
& Acc. & Prec. & Recall & F1 \\ \midrule
Random Guess & 49.91 & 55.34 & 49.65 & 52.34  \\
\hdashline
\noalign{\vskip 3pt}
bert (title+abs) & 68.33 & 67.69 & 48.54 & 56.25 \\
bert-large (title+abs) & 67.95 & 67.23 & \textbf{51.57} & 56.49 \\ 
TF-1l-8h (AC+abs) & \textbf{70.45} & \textbf{70.60} & 50.52 & \textbf{58.87} \\
\hdashline
\noalign{\vskip 3pt}
Deepseek-V3.2 & 51.52 & 55.81 & 64.86 & 60.00  \\
ChatGLM-4.7 & 57.54 & 60.02 & 61.10 & 60.56  \\
GPT-4o & 52.03 & 54.59 & 64.45 & 59.11  \\
Gemini-3-preview & 58.13 & 59.88 & \textbf{67.86} & \textbf{63.62}  \\
Claude-sonnet-4.5 & 52.67 & 56.02 & 56.14 & 56.08  \\
bert (ACI) & 64.32 & 61.72 & 41.96 & 49.50 \\
bert-large (ACI) & 64.55 & 62.89 & 40.20 & 48.46 \\
R1    & 64.64 & 62.42 & 40.72 & 48.83 \\
Gated Ave.    & 64.48 & 61.60 & 42.85 & 49.99 \\
SA-1 & \textbf{65.29} & \textbf{64.26} & 39.18 & 48.40 \\
SA-2 & 65.21 & 63.40 & 40.63 & 49.25 \\
TF-1l-1h & 64.64 & 60.34 & 45.89 & 52.00 \\
TF-1l-8h & 64.74 & 60.19 & 46.67 & 52.57 \\
\bottomrule
\end{tabular}
\label{tab:7}
\end{table}

\subsection{The results of capability prediction model}

As is discussed in Section~\ref{Sec:4.2}, we finetune two-level architecture to predict the capability representation (given by pre-trained capability-based model) with the author information and idea. We apply an AdamW optimizer with a learning rate of 0.00002 without weight decay. Next, we make comparison of: 
(a) 3-way bert model (author+idea+capability) with merging mechanism by 2-layer-8-head transformer (see Figure~\ref{Fig:S2}); (b) two-way bert model (author+capability) with merging mechanism by 2-layer-8-head transformer; (c) replace the capability model in 3-way bert model with bi-level capability prediction architecture without pretraining; (d) replace the capability model in 3-way bert model  with bi-level architecture with pretraining as in Section~\ref{Sec:4.2}. In all settings we train the model for 8 epochs. An AdamW optimizer with the same setting as in Section~\ref{Sec:5.2} and an effective batch size of 64 are applied. 

\begin{table}[htbp]
  \centering
  \small
  \caption{Results of capability model replacement}
  \begin{tabular}{lcccccc}
    \toprule
    \multirow{2}{*}{Model} & \multicolumn{2}{c}{Rating} & \multicolumn{4}{c}{Acceptance} \\
    \cmidrule(lr){2-3} \cmidrule(lr){4-7}
     & MSE & MAE & Acc & P & R & F1 \\
    \midrule
    3-way & 1.033 & 0.799 & 64.6 & 60.3 & 45.9 & 52.0 \\
    2-way & 1.076 & 0.817 & 61.7 & 62.9 & 20.9 & 31.3 \\
    w/o pre & 1.047 & 0.800 & 61.8 & 58.4 & 30.9 & 40.4 \\
    w/ pre & 1.039 & 0.798 & 62.3 & 59.1 & 32.6 & 42.0 \\
    \bottomrule
  \end{tabular}
  \label{tab:cap_rep}
\end{table}

Table ~\ref{tab:cap_rep} evaluates the impact of replacing explicit capability descriptions with predicted capability representations. The three-way model using predicted capabilities (pretrain) achieves competitive rating prediction (MSE: 1.0392, MAE: 0.7980) and acceptance prediction (Acc: 62.32\%, F1: 42.02\%), closely matching the performance of models using directly extracted capabilities (3-way-basic). This demonstrates the feasibility of inferring author capability from author information and research ideas, enabling early-stage prediction even when explicit capability descriptions are unavailable, with a minor drop in performance. 

\section{Discussion}

\subsection{Comparison with related works}

Compared with \citet{Feng2025GraphEval}, which emphasizes the goodness of research idea and its impact on the final acceptance decision, we argue that capability and author information are also important factors on the review outcomes for early stage prediction. Meanwhile, we use the full-set of the open-source papers of ICLR2024/2025 and NeurIPS2023/2024 for model training and evaluation, and consider both acceptance decision and average ratings. As we combine two conferences over two years, our task is more challenging.

\subsection{Idea and group recommendation}

The proposed model can be applied for idea and author group recommendation. This is achieved by searching for the research idea or research plan $p$ in the candidate set $P$ that gives the highest predicted average rating or acceptance odds $f(\mathbf{p}, \mathbf{a})$, given the author information and/or author capabilities $\mathbf{a}$. Meanwhile, given a research idea or a research plan $\mathbf{p}$, the trained joint model can be utilized to search for the optimal author combinations $\mathbf{a}^*$ based on textual information. Thus we have:
\begin{equation}
\begin{split}
&a^* = \text{argmax}_{a\in A} f(\mathbf{p}, \mathbf{a})\\
&p^* = \text{argmax}_{p\in P} f(\mathbf{p}, \mathbf{a}) 
\end{split}
\end{equation}

\section{Conclusion}

In this work, we introduced a capability-aware framework for early-stage research outcome prediction, leveraging only author information and research ideas, without reliance on the full text of the manuscript or experimental results. Our approach centers on a three-way transformer architecture that effectively integrates author identity, capability representations, and research ideas through different fusion mechanisms. Experimental results on a large-scale dataset from top-tier AI conferences demonstrate that our model significantly outperforms strong BERT-based baselines, achieving competitive performance in both rating and acceptance prediction. Notably, we show that explicitly modeling author capability derived from historical research outputs enhances predictive accuracy. The proposed method not only facilitates early-stage decision-making for researchers and funding bodies but also enables optimal matching between research ideas and author groups to maximize the likelihood of success.



\section{Limitations}

Despite promising results, our work has several limitations. First, our current capability model assumes static author profiles, whereas in reality, researcher skills and resources evolve over time. Second, we do not account for variations in effort or collaboration dynamics among authors with similar capability levels, which may influence final outcomes. Finally, while our dataset is substantial, it is limited to open-access submissions from two AI conferences, which may not fully represent the broader scientific landscape. Addressing these aspects in future work could further enhance the robustness and applicability of the proposed framework.

\bibliography{Reference}

\newpage
\appendix

\section{Ethical discussion} \label{Sec:A}

We acknowledge the ethical concerns associated with the use of author-related features for outcome prediction, which may risk reinforcing existing structural inequalities. However, this model is designed to assist researchers in preliminarily evaluating the potential of their research ideas during the early stages of a project, thereby saving time/cost and improving workflow efficiency. The model can be applied to match research teams with ideas and provide objective success probability estimates. In fairness-sensitive funding evaluation processes, similar black-box models should be used with caution to avoid exacerbating structural inequities. That said, in such a context, the potential bias introduced by the model itself is not greater than that of conventional large language models. In the peer-review scenario, the model is not applicable because it processes only research ideas and does not generate review comments. Furthermore, the proposed method cannot function in double-blind review settings, and consequently cannot introduce bias— where author information is withheld. Thus, risks and concerns are tied to the context and manner of use, rather than the model research itself, and such risks can be mitigated through institutional guidelines and governance. 

Indeed, one major goal of this work is not to diagnostically model the multifaceted reality of peer review, where author capability, institutional context, and idea quality interact in complex ways. By explicitly quantifying these factors, our model helps surface and measure potential systemic biases (e.g., the correlation between institutional prestige and acceptance rates), thereby providing an empirical basis for designing fairer evaluation mechanisms.

\section{Extra feature fusion strategies} \label{Sec:A.1}

\subsection{Gated Feature Fusion}

In the Gated Feature Fusion configuration (denoted as ``Gated ave.'' in Table~\ref{tab:6} and Table~\ref{tab:7}), we implement a gated fusion mechanism to dynamically integrate information from author, caption, and idea representations. The model first projects each input representation into a shared latent space through dedicated linear transformations:
\begin{equation*}
\begin{aligned}
\mathbf{x}_a &= \mathbf{W}_{a}\mathbf{h}_{author}\\ 
\mathbf{x}_c &= \mathbf{W}_{c}\mathbf{h}_{cap}\\ 
\mathbf{x}_i &= \mathbf{W}_{i}\mathbf{h}_{idea} 
\end{aligned}
\end{equation*}
where $\mathbf{h}_{author}$, $\mathbf{h}_{cap}$, and $\mathbf{h}_{idea}$ denote the original author, caption, and idea vectors respectively, extracted from their corresponding transformer encoders.

The fusion mechanism employs exponential activations and softmax normalization to compute gating weights:
\begin{equation*}
\begin{split}
\mathbf{e}_1 &= \exp(\mathbf{x}_a \odot \mathbf{x}_i) \\
\mathbf{e}_2 &= \exp(\mathbf{x}_c \odot \mathbf{x}_i) \\
\mathbf{e}_3 &= \exp(\mathbf{x}_i \odot \mathbf{x}_i) \\
\mathbf{p}_1 &= \frac{\mathbf{e}_1}{\mathbf{e}_1 + \mathbf{e}_2 + \mathbf{e}_3} \\
\mathbf{p}_2 &= \frac{\mathbf{e}_2}{\mathbf{e}_1 + \mathbf{e}_2 + \mathbf{e}_3} \\
\mathbf{p}_3 &= \frac{\mathbf{e}_3}{\mathbf{e}_1 + \mathbf{e}_2 + \mathbf{e}_3}
\end{split}
\end{equation*}
where $\odot$ represents element-wise multiplication. The final output is obtained through a weighted combination and summation:
\begin{equation*}
\mathbf{out} = \sum^d_{i=1} \left( \mathbf{p}_1 \odot \mathbf{x}_a + \mathbf{p}_2 \odot \mathbf{x}_c + \mathbf{p}_3 \odot \mathbf{x}_i \right),
\end{equation*}
where the elements of the weighted combination vector are summed up for the final output. This design enables the model to adaptively emphasize the most relevant features from each modality through learnable gating coefficients, facilitating effective interaction while maintaining representation integrity.

\begin{table*}[htbp]
  \centering{\footnotesize
  \caption{Correlation between rating prediction models and ratings}
    \begin{tabular}{crrrrrrrrrr}
    \toprule
          & \multicolumn{1}{c}{\textbf{1}} & \multicolumn{1}{c}{\textbf{2}} & \multicolumn{1}{c}{\textbf{3}} & \multicolumn{1}{c}{\textbf{4}} & \multicolumn{1}{c}{\textbf{5}} & \multicolumn{1}{c}{\textbf{6}} & \multicolumn{1}{c}{\textbf{7}} & \multicolumn{1}{c}{\textbf{8}} & \multicolumn{1}{c}{\textbf{9}} & \multicolumn{1}{c}{\textbf{10}} \\
    \midrule
    1.rating & 1.000  &       &       &       &       &       &       &       &       &  \\
    2.title & 0.267  & 1.000  &       &       &       &       &       &       &       &  \\
    3.abs & 0.383  & 0.662  & 1.000  &       &       &       &       &       &       &  \\
    4.title\_abs & 0.358  & 0.719  & 0.883  & 1.000  &       &       &       &       &       &  \\
    5.author & 0.322  & 0.721  & 0.666  & 0.710  & 1.000  &       &       &       &       &  \\
    6.cap & 0.343  & 0.716  & 0.735  & 0.738  & 0.762  & 1.000  &       &       &       &  \\
    7.idea & 0.323  & 0.727  & 0.723  & 0.754  & 0.757  & 0.801  & 1.000  &       &       &  \\
    8.author+idea & 0.366  & 0.717  & 0.722  & 0.738  & 0.876  & 0.804  & 0.894  & 1.000  &       &  \\
    9.cap+idea & 0.352  & 0.756  & 0.759  & 0.769  & 0.796  & 0.953  & 0.896  & 0.877  & 1.000  &  \\
    10.author+cap+idea & 0.381  & 0.633  & 0.670  & 0.678  & 0.795  & 0.858  & 0.777  & 0.861  & 0.857  & 1.000  \\
    \bottomrule
    \end{tabular}%
  \label{tab:rating_corr}}%
\end{table*}%

\section{Correlation of outputs from different models} \label{Sec:A.2}

The correlations between rating prediction models and ratings are shown in Table~\ref{tab:rating_corr}. Among author, capability and idea, capability has the highest correlation of 0.343 with the rating label, while author and idea's correlation with rating is close to each other. The output correlation between these three model is between 0.7 to 0.8. Author+Idea model has higher correlation with the labels than cap+idea model, while author+cap+idea model has the highest correlation with the labels.

The correlations between acceptance prediction models and acceptance labels are shown in Table ~\ref{tab:acc_corr}. There is no very high pair-wise correlation between the variables. Compared with author and ideas, the model based on solely on capability information achieves the highest correlation with the label.

\begin{table*}[htbp]
  \centering{\footnotesize
  \caption{Correlation between acceptance prediction models and acceptance labels}
    \begin{tabular}{crrrrrrrrrr}
    \toprule
          & \multicolumn{1}{c}{\textbf{1}} & \multicolumn{1}{c}{\textbf{2}} & \multicolumn{1}{c}{\textbf{3}} & \multicolumn{1}{c}{\textbf{4}} & \multicolumn{1}{c}{\textbf{5}} & \multicolumn{1}{c}{\textbf{6}} & \multicolumn{1}{c}{\textbf{7}} & \multicolumn{1}{c}{\textbf{8}} & \multicolumn{1}{c}{\textbf{9}} & \multicolumn{1}{c}{\textbf{10}} \\
    \midrule
    1.acceptance & 1.000  &       &       &       &       &       &       &       &       &  \\
    2.title & 0.093  & 1.000  &       &       &       &       &       &       &       &  \\
    3.abs & 0.374  & 0.165  & 1.000  &       &       &       &       &       &       &  \\
    4.title+abs & 0.340  & 0.138  & 0.640  & 1.000  &       &       &       &       &       &  \\
    5.author & 0.135  & 0.121  & 0.207  & 0.154  & 1.000  &       &       &       &       &  \\
    6.cap & 0.314  & 0.179  & 0.358  & 0.338  & 0.234  & 1.000  &       &       &       &  \\
    7.idea & 0.161  & 0.160  & 0.243  & 0.225  & 0.129  & 0.311  & 1.000  &       &       &  \\
    8.author+idea & 0.219  & 0.146  & 0.258  & 0.234  & 0.445  & 0.386  & 0.390  & 1.000  &       &  \\
    9.cap+idea & 0.215  & 0.181  & 0.257  & 0.236  & 0.185  & 0.531  & 0.430  & 0.397  & 1.000  &  \\
    10.author+cap+idea & 0.286  & 0.179  & 0.272  & 0.248  & 0.424  & 0.554  & 0.327  & 0.527  & 0.475  & 1.000  \\
    \bottomrule
    \end{tabular}%
  \label{tab:acc_corr}}%
\end{table*}%

\section{Model parameter comparison}
We conduct a comparative analysis of the parameter counts across the models employed in our experiments to provide insights into their complexity and scalability. Overall we summarize and compare the model parameters in Table~\ref{tab:parameter_comparison}. The baseline models, BERT-base and BERT-large, contain approximately 110 million and 340 million parameters, respectively. Our proposed three-way fusion architecture incorporates three independent transformer encoders for processing author information, capability, and research ideas. When using BERT-base as the backbone for each encoder, the total parameter count amounts to roughly 330 million, excluding the lightweight fusion modules. The fusion mechanisms themselves, including self-attention (SA1, SA2), residual linear fusion (R1) and transformer-based fusion (TF), introduce only a minimal number of additional parameters (e.g., on the order of $10^5$ to $10^6$), making them highly efficient in terms of parameter overhead.

\begin{table}[h]
\centering{\small
\caption{Comparative analysis of model parameter counts. The 'Early-Stage Prediction' column indicates whether the model can make predictions without a full manuscript. The fusion modules (SA, R1, TF) add a negligible number of parameters (<1M) and are not included in the total.}
\label{tab:parameter_comparison}
\resizebox{\linewidth}{!}{%
\begin{tabular}{llr}
\toprule
\textbf{Model} & \textbf{Components} & \textbf{Params} \\
\midrule
BERT-base & Single encoder & 110 M \\
BERT-large & Single encoder & 340 M \\
Three-way & 3$\times$BERT-base & 330 M \\
Cap.~Predictor & 2$\times$BERT-base & 220 M \\
Full framework & 3$\times$BERT-base + Cap.~Predictor & 440 M \\
\bottomrule
\end{tabular}}%
}
\end{table}

For the capability prediction module, the two-level transformer architecture introduces additional parameters corresponding to the second-level encoder and projection layers. This module adds approximately 110 million parameters when using a BERT-base sized model, bringing the total parameter count of the full capability-aware system to around 440 million in the most comprehensive configuration. Despite this increase, our model remains more parameter-efficient than naive approaches that concatenate all inputs into a single BERT-large model, while simultaneously enabling more flexible and interpretable feature interactions.

In summary, while our three-way model with capability prediction entails a higher parameter count than single-encoder baselines, it remains computationally feasible and offers superior performance without resorting to excessively large monolithic architectures.

\section{Extracted capability information}

\subsection{Examples}
In Section~\ref{Sec:4.4}, we provide the designed prompts for extracting the research ideas and author capabilities. Here we provide some examples (the author names are encoded):\\

\noindent \textbf{Case 1:} IDGen: Item Discrimination Induced Prompt Generation for LLM Evaluation

\noindent \textbf{Main Idea:} The paper proposes IDGen, a framework for generating discriminative evaluation prompts for LLMs by incorporating Item Discrimination theory, using instruction/response gradients and self-correction mechanisms to enhance question difficulty and discrimination, with innovations in automated discrimination/difficulty estimation models.

\noindent \textbf{Author Information:} \#A9F3B2 ( ms student, southeast university, ); \#7C1D8E ( researcher, tencent big data, cn ); \#E4F0C5 ( researcher, tencent big data, cn ); \#2B8D5A ( researcher, amazon, us ); \#D16B3F ( researcher, tencent ai lab, cn ); \#93CA7D ( associate professor, southeast university, cn )

\noindent \textbf{Author Capability:} The authors' capability is high in mathematical derivation, medium in theoretical analysis/proving, high in model/architecture design, very high in data collection, high in experimental design, and high in paper presentation. Core expertise includes LLM evaluation (high), educational assessment (medium), prompt engineering (high), data synthesis (high), and statistical analysis (medium). Computational work utilized ~20 A100 GPUs for 2 months. Estimated costs: a) \$50K USD; b) ~12 PhD-equivalent months."
\

\noindent \textbf{Case 2:} How DNNs break the Curse of Dimensionality: Compositionality and Symmetry Learning\

\noindent \textbf{Main Idea:} The paper investigates how deep neural networks (DNNs) overcome the curse of dimensionality by learning compositional functions and symmetries, proposing a theoretical framework using F1-norms and Sobolev spaces to derive generalization bounds for accordion networks.

\noindent \textbf{Author Information:} \#F85C29 ( assistant professor, nyu, new york university, us ); \#1B9E6F ( ms student, new york university, us ); \#C38BDC ( phd student, new york university, us )

\noindent \textbf{Author Capability:} The authors' capability is very high in mathematical derivation, very high in theoretical analysis/proving, high in model/architecture design, medium in data collection, high in experimental design, and high in paper presentation. Core expertise includes functional analysis (very high), approximation theory (very high), neural network theory (high), optimization (high), and statistical learning (high). Computational work utilized ~4 AMD Opteron 6136 CPUs for synthetic data and real-world experiments. Estimated costs: a) \$10K USD; b) ~18 PhD-equivalent months.\

\noindent \textbf{Case 3:} Decomposing and Interpreting Image Representations via Text in ViTs Beyond CLIP

\noindent \textbf{Main Idea:} The paper proposes a framework to interpret vision transformers (ViTs) by decomposing their representations into component contributions (REPDECOMPOSE) and aligning them to CLIP space for text-based interpretation (COMPALIGN), addressing the challenge of understanding feature encoding in non-CLIP ViTs. The authors introduce a scoring function to rank components by feature importance, enabling applications like image retrieval and spurious correlation mitigation.

\noindent \textbf{Author Information:} \#4D2A91 ( phd student, university of maryland, college park, us ); \#A5E62B ( assistant professor, university of maryland, college park, ); \#8F0C73 ( phd student, university of maryland, college park, )

\noindent \textbf{Author Capability:} The authors' capability is high in mathematical derivation, high in theoretical analysis/proving, high in model/architecture design, medium in data collection, high in experimental design, and high in paper presentation. Core expertise includes transformer interpretability (high), linear algebra (high), representation learning (high), computer vision (high), and algorithmic design (high). Computational work utilized ~4 RTX A5000 GPUs for weeks. Estimated costs: a) \$10K USD; b) ~12 PhD-equivalent months."\

\noindent \textbf{Case 4:} Gated Delta Networks: Improving Mamba2 with Delta Rule

\noindent \textbf{Main Idea:} The paper proposes Gated Delta Networks, combining gating mechanisms with delta rules to enhance Mamba2's memory management for improved performance in language modeling and retrieval tasks. The innovation lies in integrating adaptive memory clearance (via gating) with precise key-value updates (via delta rules), implemented through a hardware-efficient parallel algorithm.

\noindent \textbf{Author Information:} \#6BE491 ( research scientist, nvidia, us ); \#D3A07F ( vp research, nvidia, us ); \#297FC5 ( phd student, massachusetts institute of technology, us )

\noindent \textbf{Author Capability:} The authors' capability is very high in mathematical derivation, high in theoretical analysis, very high in model design, medium in data collection, high in experimental design, and high in paper presentation. Core expertise includes linear algebra (very high), RNN architectures (very high), GPU optimization (high), language modeling (high), and attention mechanisms (high). Computational work utilized ~32 H100 GPUs for 2-4 weeks. Estimated costs: a) \$50K USD; b) ~6 PhD-equivalent months."

\noindent \textbf{Case 5:} Exploring channel distinguishability in local neighborhoods of the model space in quantum neural networks

\noindent \textbf{Main Idea:} This paper investigates the trainability challenges of Quantum Neural Networks (QNNs) by analyzing ansatz architectures through quantum channel distinguishability metrics, proposing an upper bound on parameter perturbation sensitivity and validating it empirically.

\noindent \textbf{Author Information:} \#5F8C31 ( full professor, technische universitat wien, ); \#B4D269 ( phd student, technische universitat wien, at ); \#E87A9C ( phd student, forschungszentrum juelich gmbh, ); \#0C3F8A ( postdoc, technische universitat wien, at )

\noindent \textbf{Author Capability:} The authors' capability is high in mathematical derivation, high in theoretical analysis/proving, medium in model/architecture design, medium in data collection, high in experimental design, and high in paper presentation. Core expertise includes quantum information theory (high), quantum machine learning (high), numerical optimization (medium), statistical analysis (medium), and scientific writing (high). Computational work utilized ~5 NVIDIA V100 GPUs for 2 weeks. Estimated costs: a) \$10K USD; b) ~6 PhD-equivalent months."

\begin{table*}[tbp]
  \centering{\small
  \caption{Correlation of Capability Labeling}
    \begin{tabular}{ccccccc}
    \toprule
    LLM & \textbf{human-1} & \textbf{human-2} & \textbf{DS-R1} & \textbf{Claude} & \textbf{Genimi} & \textbf{GPT-5} \\
    \midrule
    \textbf{human-1} & 1.000  & 0.350  & 0.582  & 0.537  & 0.613  & 0.559  \\
    \textbf{human-2} & 0.350  & 1.000  & 0.387  & 0.325  & 0.473  & 0.405  \\
    \textbf{Deepseek-R1} & 0.582  & 0.387  & 1.000  & 0.648  & 0.598  & 0.631  \\
    \textbf{Claude-4.5} & 0.537  & 0.325  & 0.648  & 1.000  & 0.614  & 0.696  \\
    \textbf{Genimi-3} & 0.613  & 0.473  & 0.598  & 0.614  & 1.000  & 0.707  \\
    \textbf{GPT-5.2} & 0.559  & 0.405  & 0.631  & 0.696  & 0.707  & 1.000  \\
    \bottomrule
    \end{tabular}%
  \label{tab:cap_corr}}%
\end{table*}%

\subsection{Validation}



To assess the quality of the extracted author capability profiles, we performed a two-part validation. First, we manually verified the extracted research ideas and capability descriptions against the original full texts of approximately 30 randomly sampled papers. All core expertise identified by the LLM was consistently reflected in the corresponding manuscripts, confirming the accuracy and content alignment of the extracted profiles.

Second, to evaluate the reliability of the quantified capability levels, we conducted an independent annotation study. For a separate set of 30 randomly sampled manuscripts, two machine learning experts independently rated the author groups’ proficiency across five key dimensions: Mathematical Derivation, Theoretical Analysis/Proving, Model/Architecture Design, Data Collection, and Experimental Design. Each dimension was scored on a 4-point scale (1: Low, 2: Medium, 3: High, 4: Very High). We compared these human annotations with the capability levels generated automatically by several state-of-the-art LLMs: Deepseek-R1, Claude-Sonnet-4.5, Gemini-3, and GPT-5.2.

The inter-annotator correlations are presented in Table~\ref{tab:cap_corr}. Notably, the capability profiles generated by Deepseek-R1 achieve a correlation of 0.58 with Expert 1 and 0.39 with Expert 2. This inter-correlation is higher than that between the two human experts themselves (0.35), suggesting that the LLM-derived capability assessments can exhibit consistency comparable to, or even surpassing, human judgment in this task. Among the LLM evaluators, Gemini-3 shows the strongest alignment with both human experts, with correlations of 0.61 and 0.47, respectively.

These results indicate that our prompt-based methodology for extracting and quantifying author capabilities produces reliable and consistent assessments, providing a robust foundation for subsequent predictive modeling.

\section{LLM baseline setting} \label{Sec:E}

In Section~\ref{Sec:4.4}, we establish LLM-based baselines for predicting paper ratings and acceptance decisions. All candidate models are evaluated using a consistent prompt template as described below. For each instance, the research idea, extracted author capabilities, author metadata, and target venue are combined into a single formatted input string before being passed to the prompt template. To align LLM-generated average ratings with human reviewer scores, we apply linear transformation to calibrate their mean and variance to match the distribution of real ratings. For acceptance prediction, the LLMs output a predicted acceptance probability for each manuscript. We then threshold these probabilities based on the overall acceptance rate observed in our collected dataset to assign final accept/reject labels.
\begin{figure*}
\begin{tcolorbox}[
    colback=white, 
    colframe=blue!50!white, 
    colbacktitle=blue!15!white, 
    coltitle=black,      
    halign title=center,   
    title=Prompt 3:Prompt of LLM idea evaluator,
    fonttitle=\bfseries,
    boxrule=1pt,
    arc=3pt,
    width=\textwidth, 
    left=0pt,        
    right=0pt        
]
You are a senior area chair for top-tier machine learning conferences (such as ICLR or NeurIPS), with expertise in accurately assessing the quality, novelty, and completeness of submitted papers.

**Task:**
Based on comprehensive data about a paper—including author information, author capabilities, research idea, team composition, and target conference—provided by the user, conduct a quantitative evaluation. After evaluation, **output only** a Python dictionary in the format:
\{"acc\_chance": a float between 0 and 1, "rating\_ave": a float between 0 and 10\}.

**Input Format:**
The user will provide a dictionary containing the following keys:\\
- author\_cap: (String) Describes the team's proficiency in various dimensions (e.g., mathematical derivation, theoretical proof, experimental design), along with their core expertise, computational resources, and estimated costs.\\
- idea: (String) A brief description of the paper's research idea.\\
- authors: (List) A list of authors, each with details such as name, institution, and position.\\
- venue: (String) The target conference for submission (e.g., 'NeurIPS2024', 'ICLR2025').\\
**Evaluation Steps and Guidelines:**
Perform a comprehensive analysis strictly following these steps and considerations:

1.  **Analyze the Alignment Between Author Capabilities and Research Idea:**\\
    - Examine whether the core expertise listed in ``author\_cap'' (e.g., "theoretical analysis," "fairness," "optimization") directly supports the research described in ``idea'' (e.g., "theoretical bounds," "fairness violation control").\\
    - Assess whether any noted weaknesses in ``author\_cap'' (e.g., "medium capability in model design") could become a critical obstacle in executing the proposed ``idea''.

2.  **Evaluate Team Structure and Resources:**\\
    - Analyze the ``authors'' list: Is there senior researcher (professor, research scientist) supervision? Is the team composition reasonable (e.g., a PhD student working under a professor's guidance)?\\
    - Based on the computational resources (GPU count, duration) and cost estimates in ``author\_cap'', judge whether the research has sufficient experimental support to meet the standards of a top-tier conference.

3.  **Assess the Potential of the Research Idea and Conference Fit:**\\
    - Analyze whether the ``idea'' addresses an important, cutting-edge, or unresolved problem in the current field.\\
    - Determine the primary type of contribution (theoretical innovation, novel method, large-scale empirical study) and whether it aligns with the typical preferences of the target ``venue'' (ICLR or NeurIPS).\\
    - Consider the competitiveness and commonality of the topic at the target conference.

4.  **Generate Quantitative Scores:**\\
    - **``acc\_chance'' (Acceptance Probability, 0-1):** Based on the above analysis, estimate the probability of the paper being accepted. Consider: extremely high alignment and ample resources (+), novel idea with theoretical depth (+), team with inferred strong publication record (from positions and institutions, +), but also account for the generally low acceptance rates (often below 25\%) of top conferences (-). Provide a comprehensive and conservative probability estimate.\\
    - **``rating\_ave'' (Expected Average Score, 0-10):** Assuming the paper is executed fully according to the described idea and capabilities, predict the average score reviewers would give (typically 6-8 is the borderline-to-accept range, >8 is outstanding). Consider: rigor of theory/experiments, writing and presentation (as mentioned in ``author\_cap''), and the potential ceiling for innovation.

**Output Requirements:**\\
- Your response **must and can only be** the final evaluation dictionary.\\
- Do **not** include any analysis process, reasoning, or additional text.\\

Now, you will evaluate based on the dictionary information provided by the user:
\end{tcolorbox}
\end{figure*}

\end{document}